\documentclass[conference]{IEEEtran}
\IEEEoverridecommandlockouts
\usepackage{amsmath,amssymb,amsfonts}
\usepackage{algorithm}
\usepackage[numbers,sort&compress]{natbib}
\usepackage{graphicx}
\usepackage{textcomp}
\usepackage{xcolor}
\usepackage{multirow}
\usepackage{algorithmicx}
\usepackage[noend]{algpseudocode}
\usepackage{url}
\def\BibTeX{{\rm B\kern-.05em{\sc i\kern-.025em b}\kern-.08em
    T\kern-.1667em\lower.7ex\hbox{E}\kern-.125emX}}

\begin{document}

\title{OrchNAS: Orchestrated Neural Architecture Search Service for Personalised Federated Edge Intelligence \\
}

\author{
\IEEEauthorblockN{
Keya Patel, Sajib Mistry, Sheik Mohammad Mostakim Fattah, Aneesh Krishna
}
\IEEEauthorblockA{
\textit{School of EECMS, Curtin University, Australia}\\
k.patel38@postgrad.curtin.edu.au,
\{sajib.mistry, sheik.fattah, a.krishna\}@curtin.edu.au
}
}
\maketitle

\begin{abstract}
We propose OrchNAS, an energy-aware, personalised, federated edge intelligence framework that leverages a Neural Architecture Search Service to automatically design service-adaptive models for heterogeneous edge environments. The framework orchestrates the architecture search process on a server-side NAS service, enabling edge services to derive personalised architectures under device-level energy, computation, and memory constraints. We introduce an energy-aware global architecture search mechanism that learns a compact global representation across heterogeneous services. We develop an energy-efficient architecture selection mechanism that enables each service to derive a personalised subnet that satisfies its resource constraints via a progressive, greedy, energy-aware pruning strategy. We propose an energy-efficient personalised model optimisation scheme that updates service-adaptive parameters while preserving global representations, where a primal–dual optimisation mechanism enforces strict energy budgets during architecture adaptation. Experiments on real-world and benchmark datasets demonstrate the effectiveness of the proposed approach.
\end{abstract}

\begin{IEEEkeywords}
Neural Architecture Search, Federated Learning as a Service, NAS Service, Energy Efficiency, Edge Intelligence, Resource-Constrained Devices
\end{IEEEkeywords}

\section{Introduction}
The rapid growth of \textit{Edge Computing} has enabled intelligent services to operate on heterogeneous edge devices, such as smartphones, Internet of Things (IoT) sensors, and autonomous vehicles \cite{b4}. These devices are used to train machine learning (ML) models for applications such as remote surveillance, health monitoring, and intelligent transportation \cite{b2}. Deploying ML models on edge devices for intelligent services is known as \textit{Edge Intelligence}. These systems rely on AI-as-a-Service (AIaaS) platforms to manage models across devices, enabling lower latency, stronger privacy, and reduced communication overhead \cite{b4}.

\textit{Federated Learning as a Service (FLaaS)} is a cloud-based service paradigm that enables heterogeneous edge services, deployed on distributed edge devices, to collaboratively train machine learning models while keeping data local. Each service performs local training using its private data and shares only model updates with a central coordination server \cite{b26}. This approach supports privacy preservation and scalable edge intelligence across IoT environments. However, statistical heterogeneity across participating services remains a key challenge. Due to non-IID data distributions, services exhibit highly diverse data characteristics, which can hinder global model convergence and degrade performance.

\textit{Personalised Federated Learning (PFL)} adapts models to each service’s local data, improving performance in heterogeneous environments \cite{b21}. It allows each service to tailor the shared model to its local data, addressing statistical heterogeneity and improving effectiveness in FLaaS. For instance, in a health monitoring system, wearable edge services personalise the federated model to reflect individual physiological patterns.

A key challenge in PFL is high \textit{energy consumption} \cite{b7}. While personalisation improves model performance, it increases computational complexity due to additional local training and model adaptation \cite{b7}. Several PFL approaches address energy consumption through encoder-decoder-based personalisation and cluster-specific model design \cite{b7,b8}. These methods improve performance and reduce energy but rely on manually predefined architectures. A \textit{manual approach} refers to designing model architectures and tuning hyperparameters based on human expertise. This process often involves iterative trial-and-error \cite{b28}. Such approaches present several limitations, including reliance on expert knowledge, time-consuming tuning, and limited generalisation across datasets and tasks \cite{b28}. Therefore, this paper investigates automated personalisation techniques to enable efficient, scalable and adaptable federated edge intelligence.

\textit{Neural Architecture Search (NAS) Service} is a cloud-based framework that automates neural architecture optimisation by \textit{orchestrating} parallel training containers \cite{b10}. It systematically proposes candidate architectures, evaluates them via parallel training, and iteratively refines the search using performance feedback, with the resulting performance metrics serving as reward signals to guide subsequent search iterations \cite{b10}. This enables scalable architecture optimisation without manual design or search infrastructure. In \textit{Edge Intelligence}, NAS Service facilitates cloud-based architecture design, evaluates models for device constraints, and delivers adaptive models to distributed services. NAS has demonstrated strong potential in automating the design of deep learning models \cite{b1, b9}. Existing studies integrate NAS into PFL to design client-specific models, improving performance and efficiency through resource-aware, structured search \cite{b1,b2}.

In \textit{Edge Intelligence}, services operate on devices with varying hardware capabilities, memory capacity, and energy constraints \cite{b2}. Consequently, accuracy-driven personalisation can increase energy usage, leading to rapid battery depletion and limited scalability \cite{b3}. For example, a complex service model that performs well on powerful devices may quickly drain the battery on resource-constrained devices, while a lightweight service model may limit performance on high-capacity devices. Since the architecture of a service model affects computational cost and energy consumption, using the same architecture across all services is inefficient. Therefore, personalisation should adapt both model parameters and architecture to match the device constraints. NAS-driven Federated training enables collaborative architecture search across distributed clients \cite{b1,b2,b3}. To the best of our knowledge, such approaches primarily focus on accuracy and search efficiency, with little attention given to energy consumption during personalised architecture adaptation. We identify three key challenges that make achieving energy efficiency more difficult in personalised federated NAS.
\begin{itemize}
    \item \textit{Energy Consumption of NAS}: NAS is energy-intensive as it requires exploring and evaluating many architectures during the search process \cite{b9}. Each candidate model needs to be partially trained and validated to estimate its performance \cite{b9}. This repeated training increases computational and memory usage, thereby increasing energy consumption. In \textit{Edge Intelligence}, NAS consumes more energy due to the distributed architecture search across multiple clients.
    \item \textit{Global-Personalised Representation Learning}: In personalised service, models learn global knowledge, while adapting to local data and resources \cite{b2}. Nevertheless, NAS introduces challenges in designing architectures that support both global and personalised components. If the global backbone is too large, it increases the computational load; if too small, it may fail to capture useful global features. Therefore, designing NAS architectures that balance accuracy and energy across service devices is a key challenge.
    \item \textit{Service-Adaptive Architecture Search}: In FLaaS, a service represents a participating edge node with distinct data distributions and resource constraints; therefore, models are tailored to each service. The integration of NAS introduces additional complexity, as architectures need to be optimised for each service under varying data distributions and resource constraints. For example, resource-constrained services, such as wearable and mobile devices, require lightweight architectures, whereas more capable services can support deeper models. However, searching and evaluating multiple architectures incurs significant computational overhead and energy consumption, particularly in resource-constrained environments. \textit{Therefore, designing energy-efficient service-adaptive architecture search remains a key challenge.}
\end{itemize}

To address these challenges, we design \textit{OrchNAS} for \textit{Personalised Federated Edge Intelligence} by leveraging \textit{NAS Service}. OrchNAS optimises energy consumption across three critical levels: the NAS search space, the global service architecture, and personalised services. First, we introduce an \textit{energy-efficient global service architecture search} method that learns a compact shared representation. The server generates candidate architectures through evolutionary mutation and selects the architecture that minimises training energy while maintaining effective global learning. Next, we develop an \textit{energy-efficient architecture selection} mechanism that enables each service to derive a personalised service under its resource constraints. This is achieved through a progressive, greedy, energy-aware pruning strategy that removes high-computational-cost operations, producing lightweight architectures. Finally, we propose an \textit{energy-efficient personalised optimisation} that updates service-adaptive parameters while preserving global representations. A primal-dual optimisation mechanism is integrated to enforce energy budgets by penalising constraint violations. This allows each service to obtain a personalised service tailored to its data distribution and resource capability while ensuring energy-efficient training. We evaluated OrchNAS across accuracy, energy efficiency, and scalability, validating its effectiveness for \textit{Personalised Federated Edge Intelligence}. The key contributions of the paper are as follows:

\begin{itemize}
    \item A novel energy-efficient OrchNAS framework leveraging NAS services for the service-adaptive architecture design.
    \item An energy-efficient global service architecture search using evolutionary mutation and energy-aware scoring to learn a global representation.
    \item A service-adaptive architecture for personalised subnets under energy, computation, and memory constraints.
    \item A primal–dual optimisation-based personalised training method that enforces energy budgets while preserving performance accuracy.
\end{itemize}

\section{Related Work}
Prior work in \textit{Service Computing} has focused on service composition, service selection, workflow optimisation, and resource-aware management, enabling efficient coordination of distributed services in edge environments \cite{b17, b18, b19, b31, b32, b33, b34}. This establishes a strong foundation for \textit{Edge Intelligence} by enabling service deployment and efficient utilisation of distributed resources \cite{b17, b21}. However, such methods treat services as fixed computational units and do not consider adaptive architecture optimisation. To address heterogeneity, recent studies incorporate personalisation mechanisms that allow services to adapt models based on local data and resource constraints \cite{b20, b21}. Nevertheless, existing service-oriented edge intelligence frameworks largely emphasise training, composition, and resource scheduling, while overlooking automated model design \cite{b17, b21, b26}.

NAS has recently emerged as a promising approach for advancing \textit{Personalised Edge Intelligence} \cite{b1, b2, b3, b4, b5, b6}. SPIDER proposes a personalised NAS framework that searches for client-specific architectures using a shared supernet with weight sharing; however, it incurs high overhead due to supernet transmission and on-device NAS \cite{b1}. Similarly, FEDPNAS introduces a personalised NAS framework for FL that jointly learns a shared base architecture and client-specific personalised components. By incorporating context-aware operator sampling and a personalisation-aware training objective, the method improves model adaptation to heterogeneous client tasks \cite{b3}. More recently, PEACHES further extends this idea by combining shared base cells with personalised cells to improve model adaptability under heterogeneous edge environments \cite{b2}. PerFedRLNAS is a reinforcement learning (RL)-based federated NAS framework that automatically discovers personalised architectures for each client. A server-side supernet samples client-specific subnets for local training, while RL updates architecture parameters based on client performance \cite{b4}. HAPFNAS is a heterogeneity-aware personalised federated NAS framework that addresses statistical and resource heterogeneity using knowledge distillation, predictor-guided evolutionary search, and heterogeneous model aggregation \cite{b5}. A Personalised Federated Stochastic Differential Equation-based NAS (PerFedSDE-NAS) performs NAS directly on client devices, thereby preserving privacy in PFL. It combines a diffusion-based architecture generator, a performance predictor for candidate selection, and an improved supernet training and aggregation strategy to accelerate convergence \cite{b6}.

Recent advances in NAS-based PFL has demonstrate significant progress. For example, SPIDER, FEDPNAS, and PEACHES enable personalised architecture adaptation across heterogeneous clients. Despite this significance, several limitations remain. Existing methods focus on improving predictive performance through personalisation and handling data or system heterogeneity, but they largely ignore the energy consumption of architecture search and training on resource-constrained edge devices \cite{b1,b2,b3,b4,b5}. Many of these frameworks \cite{b1, b6} rely on large supernets or complex NAS procedures, which introduce significant computational overhead and communication cost during federated training. For example, transmitting large supernet structures can substantially increase computation and energy usage on edge devices with limited battery capacity. Furthermore, current approaches do not explicitly incorporate device-level energy constraints into the architecture search process, leading to architectures that may achieve high accuracy but are impractical for energy-constrained edge environments. Consequently, there remains a lack of energy-aware personalised NAS frameworks that optimise model accuracy, personalisation, and device-level energy efficiency. Therefore, designing energy-efficient personalised federated edge intelligence via NAS for resource-constrained environments remains a critical open challenge.

\section{Problem Formulation}
Let us consider an edge intelligence system based on FLaaS, consisting of a NAS Service platform and a set of edge services $\mathcal{N} = \{1,2,\dots, K\}$. Each service $k \in \mathcal{N}$ operates on a device with varying compute capacity, memory, and energy constraints, and participates in collaborative learning while keeping its data local. The NAS Service operates as a cloud-based architecture search platform that orchestrates model architecture search and distributed training across services. Each service $k \in \mathcal{N}$ holds a private dataset $D_k = \{(x_{k,j}, y_{k,j})\}_{j=1}^{n_k}$ where $(x_{k,j}, y_{k,j})$ denotes the input-label pair of the $j$-th sample and $n_k = |D_k|$ represents the local dataset size. The local objective of service $k$ is defined as
\begin{equation}
f_k(w_k, a_k | D_k) =
\frac{1}{n_k}\sum_{j=1}^{n_k} 
L(x_{k,j}, y_{k,j} | w_k, a_k)
\end{equation} where $w_k$ denotes the model parameters, $a_k$ represents the neural architecture deployed on service $k$, and 
$L(\cdot)$ is the loss function.
To address statistical heterogeneity, PFL allows each service to maintain personalised model parameters
\begin{equation}
\min_{\{w_k\}} F(W)= \sum_{k=1}^{K} p_k f_k(w_k|D_k)+\zeta R(W)
\end{equation}
where $W=\{w_1,\dots,w_K\}$ denotes the set of personalized model parameters and $R(W)$ encourages knowledge sharing among services.
NAS Service optimises model parameters and architecture to adapt device constraints. Let the supernet search space be defined as $\mathcal{A}=(E, O)$, where $E$ denotes network edges and $O$ denotes candidate operations. Each service $k$ is associated with resource constraints, compute capacity $C_k$, memory capacity $M_k$, and energy budget per round $B_k$. The computational complexity and model size of architecture $a_k$ are defined as $F(a_k)=\text{FLOPs}(a_k), \qquad
P(a_k)=\text{Params}(a_k)$. The training energy consumption of service $k$ is modelled as
\begin{equation}
E_k(a_k)=\alpha_k F(a_k)T_k
\end{equation}
where $\alpha_k$ denotes the device-specific energy coefficient and $T_k$ represents the number of local training steps. The feasible architecture set for device $k$ is therefore
\begin{equation}
\mathcal{F}_k=
\left\{
a_k\in\mathcal{A}\;|\;
E_k(a_k)\le B_k,\;
F(a_k)\le C_k,\;
P(a_k)\le M_k
\right\}
\end{equation}
Edge environments are inherently dynamic, with device states of service $s_k=(C_k, M_k, B_k)$, such as compute capacity, memory capacity, and energy budget.
We adopt a NAS Service which dynamically allocates architectures to dynamically changing device states. Let $a_k=\mathcal{O}(k,\mathcal{A},s_k,\theta_s)$ where $\theta_s$ represents the parameters of the NAS Service policy. The overall NAS Service optimisation problem for personalised, edge-intelligence can therefore be formulated as
\begin{equation}
\begin{aligned}
\min_{\{w_k\},\theta_s}\quad 
& \sum_{k=1}^{K} p_k f_k(w_k,a_k|D_k) + \zeta R(W,A) \\
\text{s.t.}\quad 
& a_k=\mathcal{O}(k,\mathcal{A},s_k,\theta_s),\; \forall k\in\mathcal{N}, a_k\in\mathcal{F}_k
\end{aligned}
\end{equation}
where $A=\{a_1,\dots,a_K\}$ denotes the set of service-specific architectures allocated by the NAS Service. The objective optimises the personalised service parameters and the architecture allocation while satisfying dynamically changing device-level resource constraints.

\section{Proposed Framework}
\begin{figure*}[htbp]
\centerline{\includegraphics[width=1.0\textwidth]{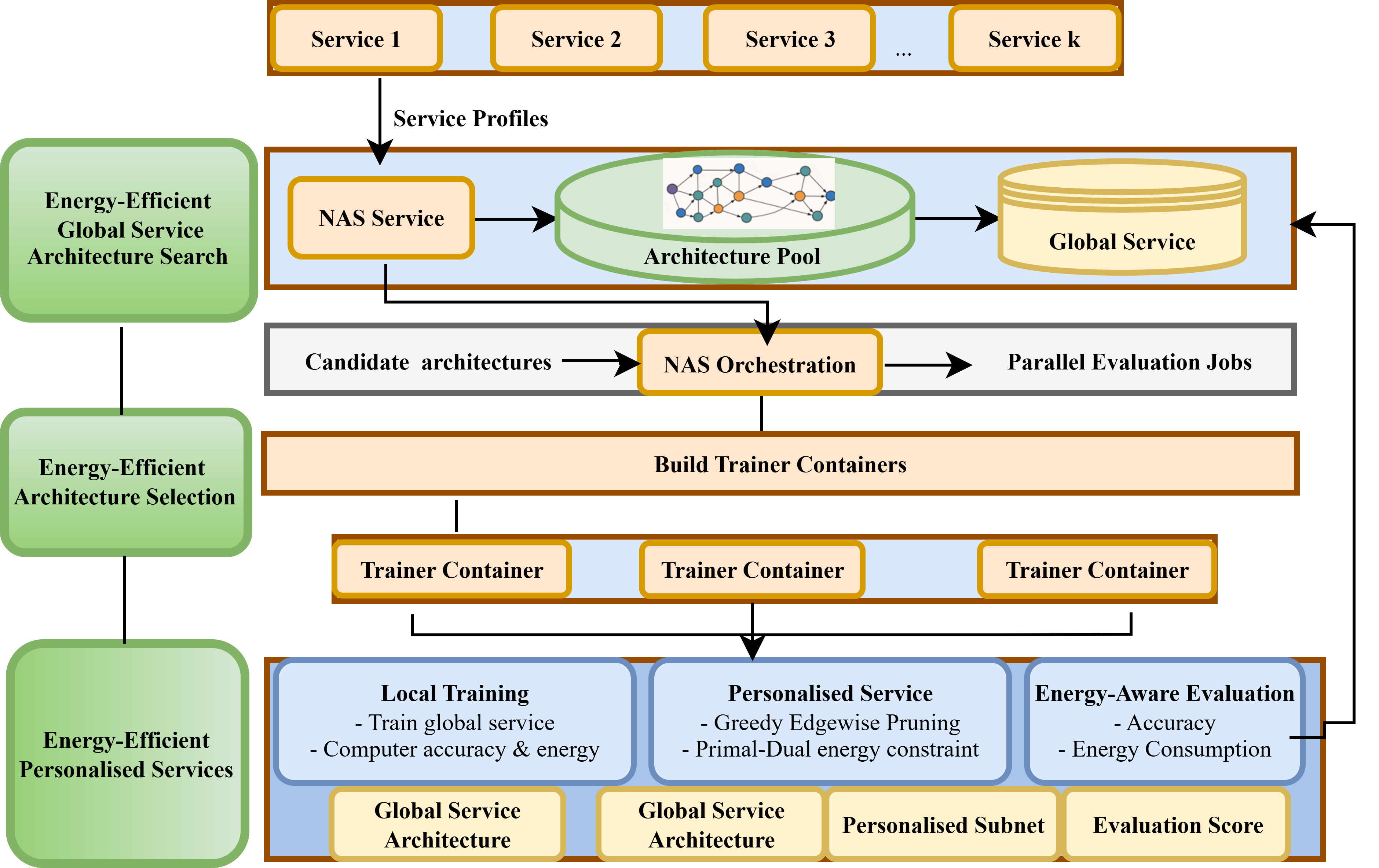}}
\caption{Overview of the OrchNAS Framework}
\label{fig:framework}
\end{figure*}

OrchNAS is an energy-aware framework that leverages NAS Service to automatically design energy-efficient, global and service-adaptive personalised services (see Fig.~\ref{fig:framework}). OrchNAS offloads the NAS process to a server-side orchestration service. The server generates and manages candidate architectures, coordinates parallel evaluation across participating services, aggregates feedback, and updates the global service, while each service derives a personalised subnet that satisfies its own energy, computation, and memory constraints. The complete training pipeline is shown in Algorithm~\ref{alg:orchnas}.

\begin{algorithm}[t]
\caption{OrchNAS Training Pipeline}
\label{alg:orchnas}
\small
\setlength{\tabcolsep}{3pt}
\renewcommand{\arraystretch}{1.1}
\begin{algorithmic}[1]
\State\textbf{Input:} Search space $\mathcal{A}$, Service set $\mathcal{K}$, Service constraints
$\{B_k^t, C_k^t, M_k^t, \alpha_k^t, T_k^t\}$
\State \textbf{Output:} Global Service $w^\ast$, Personalised subnets $\{a_k^\ast\}$,
Personalised parameters $\{v_k^\ast\}$
\State \textbf{Step 1: Initialization}
\State $w^0 \gets$ initialize global service parameters
\State $\mathcal{A}_b \subseteq \mathcal{A} \gets$ initialize candidate pool
\For{each service $k \in \mathcal{K}$}
    \State $v_k^0 \gets$ initialize personalised parameters
    \State $a_k^0 \subseteq \mathcal{A} \gets$ initialize subnet
    \State $\eta_k^0 \gets 0$
\EndFor
\State \textbf{Step 2: Communication and Global Service Search}
\For{$t = 1$ to $T$}
    \State NAS Service selects participating services $\mathcal{K}_t$
    \State NAS Service generates candidates $\mathcal{A}_b$ via mutation
    \State NAS orchestrator schedules parallel evaluation jobs
    \State NAS Service broadcasts global service parameters $w^t$

    \For{each service $k \in \mathcal{K}_t$ in parallel}
        \State Observe state $\{B_k^t, C_k^t, M_k^t, \alpha_k^t, T_k^t\}$
        \For{each candidate architecture $a \in \mathcal{A}_b$}
            \State $\hat{E}_k^t(a) \gets \alpha_k^t F(a) T_k^t$
            \State $S_k^t(a) \gets \mathrm{ValAcc}_k^t(a) - \lambda \hat{E}_k^t(a)$
        \EndFor
        \State Send $\{S_k^t(a)\}_{a \in \mathcal{A}_b}$ to NAS Service
    \EndFor

    \State $a^t \gets \arg\max_{a \in \mathcal{A}_b} \sum_{k \in \mathcal{K}_t} p_k^t S_k^t(a)$
    \State NAS Service broadcasts selected global architecture $a^t$

    \For{each service $k \in \mathcal{K}_t$}
        \State $w_k^{t+1} \gets w^t - \gamma \nabla_w F_k(w^t, v_k^t, a^t)$
        \State Derive personalised subnet $a_k^t$ from $a^t$
        \State \textbf{Step 3: Energy Feasibility and Greedy Pruning}
        \State Compute $F(a_k^t)$, $P(a_k^t)$, $E_k^t(a_k^t)$
       
        \If{$E_k^t(a_k^t) > B_k^t$ \textbf{or} $F(a_k^t) > C_k^t$ \textbf{or} $P(a_k^t) > M_k^t$}
            \State Progressive Greedy Energy-Aware Pruning ( Algorithm~\ref{alg:pruning})
        \EndIf
        \State \textbf{Step 4: Personalised Service Optimisation}
        \State$v_k^{t+1} \gets v_k^t - \gamma \Bigl(\nabla_{v_k}F_k(w^t, v_k^t, a_k^t) + \mu (v_k^t - w^t)\Bigr)$
        \State \textbf{Step 5: Dual Update}
        \State $\eta_k^{t+1} \gets \left[\eta_k^t + \rho \bigl(E_k^t(a_k^t) - B_k^t\bigr)\right]_+$
        \State Upload $w_k^{t+1}$ to NAS Service
    \EndFor

    \State $w^{t+1} \gets \sum_{k \in \mathcal{K}_t} p_k^t w_k^{t+1}$
\EndFor

\State \Return $(w^\ast,\; a_k^\ast,\; v_k^\ast)$
\end{algorithmic}
\end{algorithm}

\subsection{System Overview}
\paragraph{Supernet Definition and Initialisation}
To efficiently explore the architecture search space, we adopt the one-shot NAS approach \cite{b29}, where a single over-parameterised supernet represents all candidate architectures. All operations share weights within the supernet, allowing different architectures to inherit parameters without training each independently \cite{b29}. This significantly reduces computational cost compared to traditional NAS methods that train each candidate from scratch \cite{b29}. We define the Supernet $\mathcal{A}$ as a directed acyclic graph (DAG) representing the neural architecture search space, denoted as $\mathcal{A} = (\mathcal{E}, \mathcal{O})$. Here, $\mathcal{E} = \{e_1, e_2, \dots, e_E\}$ denotes the set of edges connecting intermediate feature nodes, and each edge $e \in \mathcal{E}$ is associated with a candidate operation set $\mathcal{O}_e = \{o_{e,1}, o_{e,2}, \dots, o_{e,O}\}$. This supernet, therefore, represents the complete architecture search space from which service-adaptive architectures are derived. 

At initialisation, NAS orchestration service initialises the global service architecture parameters $w^0$ and an initial candidate pool $\mathcal{A}_b \subseteq \mathcal{A}$. Each service $k \in \mathcal{K}$ initialises its personalised parameters $v_k^0$, an initial subnet $a_k^0 \subseteq \mathcal{A}$, and a dual variable $\eta_k^0 = 0$. The dual variable is introduced to enforce the service-specific energy constraint during training. Initialising $\eta_k^0 = 0$ ensures that no energy penalty is imposed at the start of optimisation, stabilising the early training phase.

At each communication round $t$, each selected service $k \in \mathcal{K}_t$ estimates its current state $\mathcal{C}_k^t = \{B_k^t, C_k^t, M_k^t, \alpha_k^t, T_k^t\}$ from local system statistics. The NAS Service then broadcasts the updated global parameters $w^t$ and the candidate architecture pool to the selected services. The personalised parameters $v_k^t$ and the subnet $a_k^t$ remain local and are not shared with the server. This preserves personalisation, reduces communication overhead, and protects service-specific information.

\subsection{Energy-Efficient Global Service Architecture Search}
To learn an energy-efficient global representation across heterogeneous services, we propose an energy-aware evolutionary search. Let $\mathcal{A}_b = \{a^{(1)}, a^{(2)}, \ldots, a^{(m)}\}$ denote the pool of candidate architectures derived from the supernet. At each communication round, the NAS Service updates the pool via evolutionary mutation. NAS Service generates new candidates by applying small structural mutations to previously selected architectures. Typical mutation operations include changing layer width, replacing operators on selected edges, or modifying local connectivity \cite{b30}. This strategy preserves previously discovered useful architectural patterns while evolving more energy-efficient alternatives.

Once candidate architectures are generated, a NAS orchestration dispatches candidate architectures to trainer containers and coordinates their parallel evaluation. This orchestration mechanism enables scalable architecture search by allowing multiple candidate architectures to be evaluated simultaneously. For each candidate architecture $a \in \mathcal{A}_b$, service $k$ estimates the per-round training energy as \begin{equation}
\hat{E}_k^t(a) = \alpha_k^t F(a) T_k^t
\label{eq:energy_estimate_dynamic}
\end{equation}
where $\alpha_k^t$ is a time-varying service-specific energy coefficient, $F(a)$ denotes the computational complexity (e.g., FLOPs) of architecture $a$ at round $T_k^t$. Based on this, each service computes an energy-aware score
\begin{equation}
S_k^t(a) = \mathrm{ValAcc}_k^t(a) - \lambda \hat{E}_k^t(a),
\label{eq:energy_aware_score_dynamic}
\end{equation}
where $\mathrm{ValAcc}_k^t(a)$ is the validation accuracy of architecture $a$ on service $k$ at round $t$, and $\lambda > 0$ controls the trade-off between performance and energy consumption. Each service sends the set of scores $\{S_k^t(a)\}_{a \in \mathcal{A}_b}$ to the NAS Service, which selects the global architecture by maximising the weighted global score
\begin{equation}
a^t = \arg\max_{a \in \mathcal{A}_b} \sum_{k \in \mathcal{K}_t} p_k^t S_k^t(a)
\label{eq:shared_selection_dynamic}
\end{equation}
where $\mathcal{K}_t$ is the set of participating services and $p_k^t$ is the aggregation weight. The NAS Service then sends the selected architecture $a^t$ with shared parameters $w^t$, to participating services. Each service performs local stochastic gradient descent (SGD) \cite{b27} updates on global parameters while keeping its personalised parameters fixed:
\begin{equation}
w_{k}^{t+1} = w^t - \gamma \nabla_{w} F_k(w^t, v_k^t, a^t)
\label{eq:shared_update_dynamic}
\end{equation}
where $\gamma$ is the learning rate and $F_k(w,v,a)$ is the local objective on service $k$. After local optimisation, the updated parameters $w_k^{t+1}$ are uploaded to the NAS Service for aggregation.

The global service architecture captures transferable features across services. By optimising architectures via energy-aware criteria \eqref{eq:energy_aware_score_dynamic}--\eqref{eq:shared_update_dynamic}, OrchNAS identifies compact architectures that preserve global knowledge while reducing computation and energy costs.
\subsection{ Energy-Efficient Architecture Selection}

\subsubsection{Dynamic Energy Modelling and Feasibility Constraints}

After the global service architecture update, each service $k$ evaluates whether its current subnet $a_k^t$ satisfies device-level resource constraints under its current state. The architectural complexity is measured in terms of computational and memory cost as $F(a_k^t) = \mathrm{FLOPs}(a_k^t), \qquad P(a_k^t) = \mathrm{Params}(a_k^t)$, where $F(a_k^t)$ denotes the number of floating-point operations and $P(a_k^t)$ denotes the number of trainable parameters (model size). The per-round training energy is modelled as
\begin{equation}
E_k^t(a_k^t) = \alpha_k^t F(a_k^t) T_k^t
\label{eq:client_energy_dynamic}
\end{equation}
where $(a_k^t)$ is a service-specific energy coefficient capturing energy cost per floating-point operations at round $t$. Each service then enforces the following time-varying feasibility conditions:
\begin{equation}
E_k^t(a_k^t) \leq B_k^t,\qquad F(a_k^t) \leq C_k^t,\qquad P(a_k^t) \leq M_k^t
\label{eq:feasibility_dynamic}
\end{equation}
where $B_k^t$ is the available energy budget, $C_k^t$ is the computational capacity, and $M_k^t$ is the memory capacity at round $t$. These constraints ensure that the selected personalised architecture is feasible on the target service without excessive energy drain or resource overload under dynamic conditions. To satisfy each constraint, each service $k$ refines its subnet $a_k^t$ through Progressive Greedy Energy-Aware Pruning. The goal of refinement is to solve a constrained local problem as
\begin{equation}
\begin{aligned}
&\min_{a_k^t} \quad & F_k(w_k^t, v_k^t, a_k^t) \\
\text{s.t.} \quad 
& E_k^t(a_k^t) \le B_k^t 
& F(a_k^t) \le C_k^t 
& P(a_k^t) \le M_k^t
\end{aligned}
\end{equation}

\subsubsection{Greedy Edge-Wise Energy-Aware Pruning}
We adopt a \emph{Progressive Greedy Energy-Aware Pruning} strategy (see Algorithm~\ref{alg:pruning}) that minimises an energy-regularised objective
\begin{equation}
\min_{a_k^t} F_k(w_k^t, v_k^t, a_k^t) + \beta E_k^t(a_k^t)
\label{eq:regularised_pruning_dynamic}
\end{equation}
where $\beta > 0$ controls the trade-off between model utility and energy usage. For each operation $o$ on edge $e$, the service evaluates the impact of removing that operation from the subnet. The accuracy degration is computed (\ref{eq:delta_acc_dynamic}) and energy reduction is computed as shown in (\ref{eq:delta_energy_dynamic})
\begin{equation}
\Delta \mathrm{Acc}_o^t = \mathrm{Acc}(a_k^t) - \mathrm{Acc}(a_k^t \setminus o)
\label{eq:delta_acc_dynamic}
\end{equation}
\begin{equation}
\Delta E_o^t = E_k^t(a_k^t) - E_k^t(a_k^t \setminus o)
\label{eq:delta_energy_dynamic}
\end{equation}
Each operator is ranked using the multi-objective score (\ref{eq:pruning_score_dynamic}). A smaller score indicates that removing the operator yields large energy savings with limited accuracy degradation. Therefore, operators with the lowest scores are pruned first. The pruning process proceeds progressively across edges until the feasibility conditions in (\eqref{eq:feasibility_dynamic}) are satisfied.
\begin{equation}
\mathrm{Score}^t(o) = \Delta \mathrm{Acc}_o^t - \beta \Delta E_o^t.
\label{eq:pruning_score_dynamic}
\end{equation}
This dynamic adaptation enables OrchNAS to respond to changes in data distributions, improving the practical deployability of personalised models in real-world edge environments (see Algorithm 2). While greedy edge-wise pruning reduces architectural complexity, it does not guarantee strict satisfaction of the energy constraint. Therefore, OrchNAS incorporates a primal-dual optimisation mechanism that continuously enforces the energy budget by penalising violations during training. These two mechanisms operate in complementary ways: pruning ensures structural feasibility, whereas primal–dual optimisation guarantees constraint satisfaction under standard saddle-point conditions. This design enables energy-efficient architecture search with reliable budget enforcement.

\begin{algorithm}[t]
\caption{Progressive Greedy Energy-Aware Pruning}
\label{alg:pruning}
\small
\setlength{\tabcolsep}{3pt}
\renewcommand{\arraystretch}{1.1}
\begin{algorithmic}[1]

\State \textbf{Input:} Supernet $\mathcal{A}=(\mathcal{E},\{\mathcal{O}_e\})$; Subnet $a_k^t$; Pruning coefficient $\beta$; Dynamic budgets $B_k^t, C_k^t, M_k^t$; Energy coefficient $\alpha_k^t$

\State \textbf{Output:} Pruned subnet $a_k^{t+1}$

\State $F(a_k^t) \leftarrow \mathrm{FLOPs}(a_k^t)$

\State $P(a_k^t) \leftarrow \mathrm{Params}(a_k^t)$

\State
$E_k^t(a_k^t) \leftarrow \alpha_k^t F(a_k^t)\tau_k^t$

\State
$\mathrm{Feasible}(a_k^t)
\Leftrightarrow
\big(E_k^t(a_k^t)\leq B_k^t\big)
\wedge
\big(F(a_k^t)\leq C_k^t\big)
\wedge
\big(P(a_k^t)\leq M_k^t\big)$

\While{$\mathrm{Feasible}(a_k^t)=\mathrm{false}$}

    \State $A_{\mathrm{base}} \leftarrow Acc_k(a_k^t)$

    \State $E_{\mathrm{base}} \leftarrow E_k^t(a_k^t)$

    \For{each edge $e \in \mathcal{E}$}

        \If{$|a_k^t(e)| \leq 1$}
            \State continue
        \EndIf

        \For{each operator $o \in a_k^t(e)$}

            \State $a' \leftarrow a_k^t \setminus \{(e,o)\}$

            \State
            $
            \Delta Acc_o^t \leftarrow A_{\mathrm{base}} - Acc_k(a')
            $

            \State 
            
            $\Delta E_o^t \leftarrow E_{\mathrm{base}} - E_k^t(a')$

            \State
            $
            Score^t(o) \leftarrow \Delta Acc_o^t - \beta \Delta E_o^t
            $

        \EndFor

        \State $o_{\min} \leftarrow \arg\min_{o \in a_k^t(e)} Score^t(o)$

        \State $a_k^t \leftarrow a_k^t \setminus \{(e,o_{\min})\}$

        \If{$\mathrm{Feasible}(a_k^t)$}
            \State break
        \EndIf

    \EndFor

    \EndWhile

\State $a_k^{t+1} \leftarrow a_k^t$, \textbf{return} $a_k^{t+1}$
\end{algorithmic}
\end{algorithm}

\subsection{Energy-Efficient Personalised Model Optimisation}
Once a feasible service-specific subnet $a_k^t$ has been obtained, service $k$ updates only its personalised parameters $v_k$, while keeping the global service architecture parameters fixed. Since the architecture is fixed at this stage, the energy term becomes constant with respect to $v_k$ and no longer appears explicitly in the weight update. The personalised optimisation problem is
\begin{equation}
\min_{v_k}
F_k(w_{\mathrm{global}}^t, v_k, a_k^t)
+
\frac{\mu}{2}\|v_k - w_{\mathrm{global}}^t\|^2
\label{eq:personalised_obj_dynamic}
\end{equation}
where $F_k(w_{\mathrm{global}}^t, v_k, a_k^t)$ optimizes predictive performance on the service $k$'s data, $\mu > 0$ controls the strength of proximal regularization,$\frac{\mu}{2}\|v_k - w_{\mathrm{global}}^t\|^2$ prevents excessive divergence from the shared representation. The personalised service's parameters are updated by SGD as
\begin{equation}
v_k^{t+1}
=
v_k^t
-
\gamma
\left(
\nabla_{v_k} F_k(w_{\mathrm{global}}^t, v_k^t, a_k^t)
+
\mu (v_k^t - w_{\mathrm{global}}^t)
\right)
\label{eq:personalised_update_dynamic}
\end{equation}
This update balances minimising local empirical risk with maintaining consistency with the global representation.

\subsubsection{Primal--Dual Energy-Constrained Architecture Personalisation}
To ensure strict satisfaction of the energy constraint beyond structural pruning, we adopt a primal-dual optimisation framework that transforms the constrained personalisation problem into a saddle-point optimisation task \cite{b16}. The personalised parameters and architectures are optimised via gradient descent, while energy constraints are enforced through projected gradient ascent on dual variables using a \textit{Lagrangian formulation}. For service $k$, the energy-constrained architecture search problem is formulated as
\begin{equation}
\min_{a_k^t} F_k(w_{\mathrm{global}}^t, v_k^t, a_k^t)
\quad
\text{s.t.}
\quad
E_k^t(a_k^t) \leq B_k^t
\label{eq:primal_problem_dynamic}
\end{equation}
where $E_k^t(a_k^t)$ denotes the per-round training energy, and $B_k^t$ is service-specific energy budget.
\paragraph{Lagrangian Formulation}
To solve the constrained problem, we introduce a non-negative dual variable $\eta_k^t \geq 0$ for the energy constraint. The corresponding Lagrangian is defined as
\begin{equation}
\mathcal{L}_k^t(a_k^t,\eta_k^t)
=
F_k(w_{\mathrm{global}}^t, v_k^t, a_k^t)
+
\eta_k^t \big(E_k^t(a_k^t)-B_k^t\big)
\label{eq:lagrangian_dynamic}
\end{equation}
The constrained optimisation problem can therefore be transformed into the following saddle-point problem, which enables the framework to optimise architecture selection while dynamically enforcing the energy constraint.
\begin{equation}
\min_{a_k^t}\max_{\eta_k^t \geq 0} \mathcal{L}_k^t(a_k^t,\eta_k^t)
\label{eq:saddle_point_dynamic}
\end{equation}
\begin{itemize}
    \item Primal Update (architecture update)
\end{itemize}
Given the current dual variable $\eta_k^t$, the architecture is updated to minimise the Lagrangian as shown in (\ref{eq:primal_update_dynamic}). In practice, this minimisation is realised through greedy pruning. A larger dual variable, $\eta_k^t$, increases the penalty for energy consumption, encouraging lower-complexity architectures when the budget is violated.
\begin{equation}
a_k^{t+1}
=
\arg\min_{a_k}
\left[
F_k(w_{\mathrm{global}}^t, v_k^t, a_k) + \eta_k^t E_k^t(a_k)
\right]
\label{eq:primal_update_dynamic}
\end{equation}
\begin{itemize}
    \item Dual Update (Constraint enforcement)
\end{itemize}
The dual variable is updated by projected gradient ascent by (\ref{eq:dual_update_dynamic}), where $\rho > 0$ is the dual step size and $[\cdot]_+ = \max(0,\cdot)$ denotes projection onto the non-negative orthant. If the energy budget is violated, $\eta_k^t$ increases, thereby strengthening the penalty on energy consumption in subsequent rounds. If the energy budget is satisfied, the dual variable stabilises. This primal-dual mechanism ensures that the energy budget is satisfied while maintaining good predictive accuracy.
\begin{equation}
\eta_k^{t+1}
=
\left[
\eta_k^t + \rho \big(E_k^t(a_k^{t+1}) - B_k^t\big)
\right]_+
\label{eq:dual_update_dynamic}
\end{equation}
\textbf{Theorem 1 (Energy Constraint Satisfaction).}
Assume that the local objective function $F_k(w, v_k, a_k)$ is convex and Lipschitz continuous, and the energy function $E_k(a_k)$ is bounded. Then, under appropriate learning rates, the above primal--dual updates ensure:
\begin{equation}
\limsup_{T \to \infty}
\frac{1}{T} \sum_{t=1}^{T}
\left(E_k(a_k^t) - B_k^t \right)
\leq \epsilon
\end{equation}
where $\epsilon = \mathcal{O}(\rho)$ is constant dependent on dual step size,$(E_k(a_k^t) - B_k^t)$ is constraint violation. The average violation over $T$ rounds is $\frac{1}{T}\sum_{t=1}^{T}$, $\limsup_{T \to \infty}$ captures long-term behaviour. The bound $\epsilon$ indicates violations remain small.

\textit{Proof}- The result follows from standard saddle-point optimisation analysis\cite{b16}. The Lagrangian is convex in $a_k$ and linear in $\eta_k$. The dual update penalises constraint violations over time, and the cumulative constraint violation is bounded by $\mathcal{O}(\sqrt{T})$. Dividing by $T$ yields asymptotic satisfaction of the energy constraint.

\subsection{Server Aggregation and Global Update}
After completing local updates, each service uploads its updated global parameters $w_k^{t+1}$ to the NAS Service. Personalised service parameters $v_k^{t+1}$ and personalised subnet structures $a_k^{t+1}$ remain local and are not transmitted. The NAS Service aggregates the global parameters using weighted averaging as follows:
\begin{equation}
w^{t+1}
=
\sum_{k \in \mathcal{K}_t} p_k^t w_k^{t+1}
\qquad
p_k^t = \frac{|D_k|}{\sum_{j \in \mathcal{K}_t} |D_j|}
\label{eq:aggregation_dynamic}
\end{equation}
where $|D_k|$ denotes the data size of service $k$. This aggregation approximates a single global gradient descent step and enables the NAS Service to learn consistent global representations across heterogeneous data. The process is repeated $T$ times. At convergence, the NAS Service obtains the final global parameters $w^\ast$, while each service retains its personalised subnet $a_k^\ast$ and parameters $v_k^\ast$. OrchNAS achieves a balanced trade-off between prediction accuracy and training energy, enabling scalable and energy-efficient personalised federated edge intelligence under dynamic service conditions.

\section{Experiment Results and Analysis}
A series of experiments is designed to evaluate the proposed OrchNAS. We evaluate the efficiency of OrchNAS by comparing it with representative PFL and NAS-based methods. We further assess the effectiveness of the global architecture search, architecture selection, and personalised optimisation components using multiple baseline strategies. To analyse scalability, we increase the number of participating services and measure accuracy and energy consumption. We conducted a sensitivity analysis for a selected hyperparameter and system configuration. All experiments are implemented in Python on a system with 16 GB RAM and 500 GB of local storage. The source code and experimental details are publicly available.\footnote{\url{https://github.com/keyadata/OrchNAS}}.

\subsection{Experiment Set Up and Datasets}
We simulate a heterogeneous edge environment with 50 edge services coordinated by a central server. In each communication round, 10 services perform local SGD in architecture searching and training with momentum 0.9. Each service has dynamic resource constraints that may vary across communication rounds. Constraints, including device-specific energy, computation, and memory budgets, guide the architecture selection process to derive feasible personalised architectures for the service. To represent heterogeneous hardware capabilities, services are classified into low-, medium-, and high-resource groups. The architecture search space is defined by backbone depth $\{2,3,5\}$, channel width $\{32,64,128,192,256\}$, activation functions $\{\mathrm{ReLU}, \mathrm{GELU}, \tanh\}$, operation types $\{\text{standard},\ \text{depthwise separable},\ \text{inverted bottleneck}\}$, and kernel sizes $\{3,5,7\}$. This results in $|\mathcal{A}| = 3 \times 5 \times 3 \times 3 \times 3 = 405$ candidate architectures.

We use publicly available datasets to evaluate the proposed OrchNAS framework, as they provide complementary environments for analysing NAS service in edge intelligence. CIFAR-10 \cite{b22}, and Street View House Numbers (SVHN) \cite{b23} enable efficient exploration of the search space under relatively lightweight settings, while CIFAR-100 \cite{b22} and Tiny-ImageNet\cite{b24} introduce greater visual complexity and class diversity, making them suitable for evaluating the robustness of personalised architecture adaptation. ImageNet-100 \cite{b25} further provides a more challenging large-scale setting to assess the scalability of OrchNAS and the effectiveness of personalised subnet optimisation. All datasets are partitioned using a non-IID Dirichlet split to simulate real-world data heterogeneity. We train all methods for 100 rounds, with 20 local epochs per service and a batch size of 64. We measure accuracy, energy consumption (J), and FLOPs for the accuracy-energy trade-offs for efficient, effective and scalable personalised edge intelligence deployment. 

\subsection{Experiment 1: Efficiency of OrchNAS}
To evaluate the efficiency of OrchNAS, we compare it with representative PFL and NAS methods. \textit{SPIDER} \cite{b1}, \textit{PerFedRLNAS} \cite{b4}, and \textit{diffusion-driven personalised NAS} \cite{b6}. These approaches explore personalised model architectures or adaptive training strategies to address data and device heterogeneity in FL environments. We include the \textit{Green-EDP} \cite{b7} and \textit{MCFL} \cite{b8} baselines, which focus on energy-aware training and client clustering for PFL. All methods are evaluated under the same experimental configuration to ensure a fair comparison. We conduct experiments on five datasets of varying complexity-CIFAR-10, CIFAR-100, SVHN, Tiny-ImageNet, and ImageNet-100 to comprehensively evaluate the performance of OrchNAS across diverse edge intelligence scenarios. For lightweight datasets (CIFAR-10 and SVHN), we use a small CNN \cite{b1} architecture to enable efficient architecture search under strict resource constraints. For more complex datasets (CIFAR-100, Tiny-ImageNet, and ImageNet-100), we adopt MobileNet-V2 [11] to reflect a realistic edge environment. We measure \textit{accuracy} as the average classification accuracy across participating services on the test dataset, while \textit{energy consumption} is estimated from the \textit{computational cost} of each architecture, accounting for its \textit{FLOPs} and local training workload under service-specific conditions. Table~\ref{tab:comparison} summarises the efficiency results across all datasets. OrchNAS outperforms the baseline methods in both accuracy and resource efficiency. On CIFAR-10, OrchNAS achieves 91.0\% accuracy (+1.1\%) and reduces energy consumption by up to 30\% and computational cost to 96M FLOPs. On ImageNet-100, it reaches 74.3\% accuracy, exceeding the strongest NAS baseline by 2.2\%, while reducing energy consumption by 19\% and computational complexity to 235M FLOPs. Similar trends are observed on CIFAR-100 and Tiny-ImageNet, where OrchNAS achieves higher accuracy with lower energy and computational cost. On SVHN, it maintains strong performance under non-IID data and reduces energy consumption, demonstrating robustness in real-world edge scenarios. These results show that OrchNAS learns lightweight, energy-efficient personalised architectures across diverse datasets, improving accuracy, while reducing energy consumption and computational cost across heterogeneous edge services.
\begin{table*}[t]
\caption{Performance comparison of OrchNAS with baseline methods across multiple datasets}
\label{tab:comparison}
\centering
\footnotesize
\renewcommand{\arraystretch}{1.1}
\setlength{\tabcolsep}{1.2pt}
\scriptsize
\begin{tabular}{l|ccc|ccc|ccc|ccc|ccc}
\hline
\textbf{Method} 
& \multicolumn{3}{c}{\textbf{CIFAR-10}} 
& \multicolumn{3}{c}{\textbf{CIFAR-100}} 
& \multicolumn{3}{c}{\textbf{SVHN}} 
& \multicolumn{3}{c}{\textbf{Tiny-ImageNet}} 
& \multicolumn{3}{c}{\textbf{ImageNet-100}} \\
\cline{2-16}
 & Acc & Energy & FLOPs 
 & Acc & Energy & FLOPs 
 & Acc & Energy & FLOPs 
 & Acc & Energy & FLOPs 
 & Acc & Energy & FLOPs \\
\hline
SPIDER\cite{b1} & 86.7 & 52.4 & 125 & 62.3 & 88.1 & 210 & 92.1 & 35.2 & 78 & 58.4 & 120.5 & 260 & 70.8 & 142.5 & 310 \\
PerFedRLNAS\cite{b4} & 85.2 & 49.8 & 129 & 60.9 & 84.3 & 215 & 91.4 & 34.1 & 82 & 56.7 & 118.2 & 268 & 69.4 & 139.7 & 322 \\
Diffusion NAS\cite{b6} & 89.9 & 48.6 & 112 & 65.1 & 80.7 & 198 & 93.5 & 32.8 & 72 & 60.3 & 110.6 & 240 & 72.1 & 130.4 & 284 \\
Green-EDP\cite{b7}& 88.1 & 41.5 & 105 & 63.4 & 70.5 & 185 & 92.6 & 29.4 & 70 & 58.9 & 102.4 & 228 & 69.3 & 118.2 & 260 \\
MCFL\cite{b8} & 87.3 & 44.7 & 110 & 61.8 & 75.2 & 192 & 92.0 & 30.8 & 75 & 57.2 & 108.7 & 235 & 68.9 & 121.6 & 291 \\
\textbf{OrchNAS} & \textbf{91.0} & \textbf{36.9} & \textbf{96} 
& \textbf{67.8} & \textbf{62.4} & \textbf{170} 
& \textbf{94.2} & \textbf{26.1} & \textbf{65} 
& \textbf{63.5} & \textbf{92.1} & \textbf{210} 
& \textbf{74.3} & \textbf{105} & \textbf{235} \\
\hline
\end{tabular}
\end{table*}

\subsection{Experiment 2: Effectiveness of OrchNAS}
\subsubsection{Evaluation of Energy-Efficient Global Architecture Search}
We evaluate the effectiveness of the proposed energy-efficient global service search mechanism on CIFAR-10 using four baselines inspired by prior NAS approaches. These include \textit{Largest backbone}, which selects the largest architecture in the search space to maximise model capacity \cite{b9}; \textit{Smallest Backbone}, which selects the smallest architecture to minimise computational cost, similar to lightweight CNN designs for resource-constrained devices \cite{b12}; \textit{Accuracy-Only Search}, which selects the architecture with the highest validation accuracy without considering energy consumption \cite{b9}; and \textit{Random Search}, which randomly samples architectures from the search space and is widely used as a baseline in NAS evaluation \cite{b13}. As shown in Table~\ref{tab:backbone}, OrchNAS achieves 85.2\% test accuracy with 1.03 J energy and 79M FLOPs. This is more effective than the Largest Backbone (1.82 J, 120M FLOPs) and Accuracy-Only Search (1.56 J, 108M FLOPs). This improvement is due to the evolutionary mutation process, which explores diverse architectures and selects architectures with higher energy-aware scores. Also, the global service learns a compact global architecture that can be reused across services. This reduces redundant architecture search during the NAS process. Overall, OrchNAS balances prediction performance and energy efficiency in edge intelligence.

\begin{table}[htbp]
\caption{Comparison of Global Architecture Search Strategies}
\begin{center}
\centering
\footnotesize
\renewcommand{\arraystretch}{1.1}
\setlength{\tabcolsep}{1.2pt}
\scriptsize
\begin{tabular}{c|c|c|c|c}
\hline
\textbf{Method} & \textit{Val. Acc.} & \textit{Test Acc.} & \textit{Energy (J)} & \textit{FLOPs (M)} \\
\hline
Largest Backbone\cite{b9} & 86.1 & 84.5 & 1.82 & 120 \\
Smallest Backbone\cite{b12} & 79.2 & 76.8 & 0.72 & 48 \\
Accuracy-Only Search\cite{b3} & 86.7 & 84.8 & 1.56 & 108 \\
Random Search\cite{b13} & 83.1 & 80.9 & 1.21 & 86 \\
\textbf{OrchNAS} & \textbf{85.9} & \textbf{85.2} & \textbf{1.03} & \textbf{79} \\
\hline
\end{tabular}
\label{tab:backbone}
\end{center}
\end{table}

\subsubsection{Evaluation of Energy-Efficient Architecture Selection}
We evaluate an energy-efficient architecture selection mechanism in which each service derives a personalised subnet from the global architecture via progressive pruning to satisfy device-level energy, computation, and memory constraints. We compare OrchNAS with several architecture adaptation strategies on CIFAR-10, inspired by prior work. No Adaptation directly deploys the global architecture without structural modification, meaning all clients use the same full model regardless of device constraints \cite{b5}. Random Pruning randomly removes network components until device constraints are satisfied, which is a common baseline in NAS evaluation \cite{b13}. FLOPs-Based Pruning removes operations according to their computational cost to reduce model complexity \cite{b9}. Uniform Scaling uniformly reduces the width of network layers to obtain a smaller model, a strategy widely used for resource-constrained devices \cite{b12}. Table \ref{tab:arch_selection} summarises the results in terms of personalised test accuracy, energy reduction, computational complexity, model size, and feasibility rate. Compared to baselines, the OrchNAS architecture selection mechanism achieves 84.2\% classification accuracy with improving resource efficiency across heterogeneous services. The selected architectures reduce training energy consumption by 35.8\%, demonstrating the effectiveness of the energy-aware selection strategy. In addition, OrchNAS identifies lightweight models with only 71M FLOPs and 0.88M parameters, which lowers computational and memory requirements for edge deployment. The method also achieves a 92.6\% feasibility rate, indicating that most selected architectures satisfy device-level resource constraints, making the approach suitable for energy-constrained edge environments. OrchNAS progressively removes edges with the lowest energy-aware contribution in the supernet, preserving important feature paths while eliminating redundant computations. As a result, low-resource devices obtain compact subnet architectures, whereas high-resource devices retain larger architectures. This approach enables OrchNAS to derive service-adaptive personalised architectures while improving energy efficiency.

\begin{table}[t]
\centering
\footnotesize
\renewcommand{\arraystretch}{1.1}
\setlength{\tabcolsep}{1.2pt}
\scriptsize
\caption{Comparison of architecture selection methods}
\label{tab:arch_selection}
\begin{tabular}{l|c|c|c|c|c}
\hline
\textbf{Method} & Acc(\%) & Energy Red. (\%) & FLOPs (M) & Params (M) & Feas. Rate (\%) \\
\hline
No Adaptation\cite{b5} & 85.4 & 0.0  & 108 & 1.42 & 41.2 \\
Random Pruning\cite{b13} & 80.1 & 24.5 & 82  & 1.01 & 76.8 \\
FLOPs Pruning \cite{b9}& 81.6 & 31.7 & 75  & 0.94 & 84.5 \\
Uniform Scaling\cite{b12} & 80.9 & 28.9 & 79  & 0.98 & 80.7 \\
\textbf{OrchNAS} & \textbf{84.2} & \textbf{35.8} & \textbf{71} & \textbf{0.88} & \textbf{92.6} \\
\hline
\end{tabular}
\end{table}

\subsubsection{Evaluation of Energy-Efficient Personalised Optimisation}
We compare the energy-efficient personalised optimisation on CIFAR-10 using three baseline variants inspired by prior work:
(i) \textit{Standard Federated Optimisation}, which performs FL 
training without personalisation; (ii) \textit{Personalised Training without Energy 
Regularisation}, which updates personalised parameters but does not enforce 
energy constraints; and (iii) \textit{Energy-Regularised Training without Dual Updates}, 
which introduces energy regularisation but does not apply the primal–dual optimisation 
mechanism. Evaluation uses multiple metrics: Average personalised accuracy (mean accuracy across all services), worst-service accuracy (accuracy of the lowest-performing service), energy violation rate (percentage of services exceeding the energy budget), and average training energy consumption (mean energy required for local training). Table~\ref{tab:optimisation} presents the results. OrchNAS achieves the highest 
personalised accuracy (85.1) and worst-service accuracy (79.3), showing performance and robustness across heterogeneous services. Also, OrchNAS significantly reduces energy budget violations to 
4.8\%, compared with 18.4\% in standard FL optimisation and 26.7\% in 
unconstrained personalised training with 
lowest average training energy consumption (1.07). This improvement is mainly due to the primal–dual optimisation mechanism. The primal update optimises personalised model parameters, and the dual update penalises energy violations. As a result, OrchNAS learns resource-aware personalised models that maintain strong accuracy while reducing energy consumption in edge intelligence.

\begin{table*}[t]
\centering
\caption{Comparison of personalised model optimisation methods.}
\label{tab:optimisation}
\small
\setlength{\tabcolsep}{4pt}
\begin{tabular}{l|c|c|c|c}
\hline
\textbf{Method} & Avg. Pers. Acc. & Worst-Client Acc. & Energy Viol. Rate (\%) & Avg. Train Energy \\
\hline
Std. FL Optimisation \cite{b26}& 80.1 & 74.2 & 18.4 & 1.62 \\
Personalised w/o Energy Reg.\cite{b3} & 84.3 & 78.1 & 26.7 & 1.74 \\
Energy Reg. w/o Dual Update \cite{b7} & 83.2 & 76.8 & 11.9 & 1.28 \\
\textbf{OrchNAS} & \textbf{85.1} & \textbf{79.3} & \textbf{4.8} & \textbf{1.07} \\
\hline
\end{tabular}
\end{table*}

\subsection{Experiment 3: Scalability Analysis of OrchNAS}
\begin{figure}[htbp]
\centerline{\includegraphics[width=0.48\textwidth]{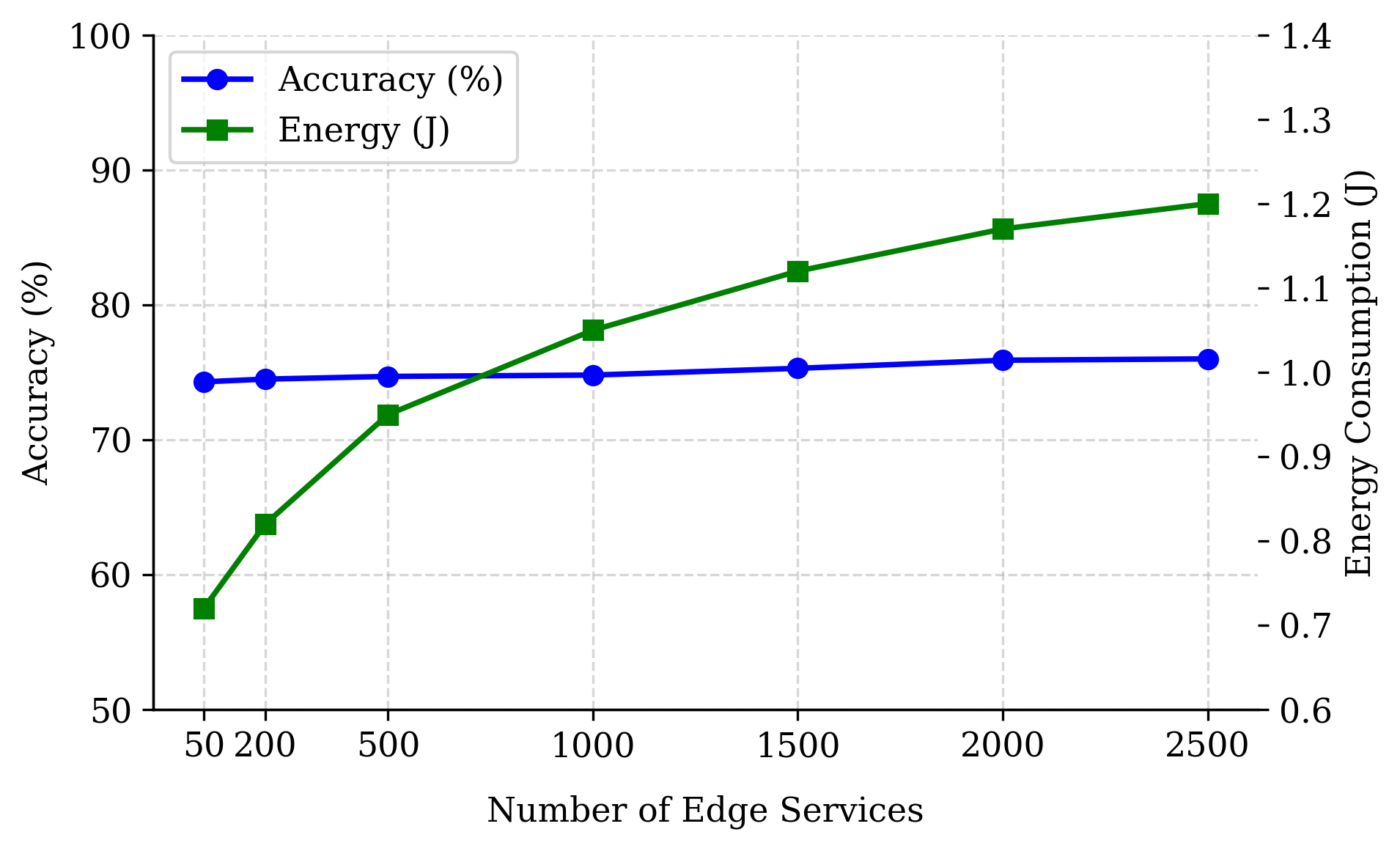}}
\caption{Scalability of OrchNAS on ImageNet-100}
\label{fig:orchnas_scalability}
\end{figure}

To evaluate the scalability of OrchNAS on ImageNet-100, we increase the number of participating edge services from 50 to 2500. The results in Fig.~\ref{fig:orchnas_scalability} show that the personalised accuracy increases from $74.3\%$ to $76.0\%$, with minor fluctuations across intermediate scales. These variations are expected in large-scale federated environments due to stochastic client participation and non-IID data distributions. Nevertheless, the overall improvement indicates that the global architecture benefits from increased data diversity, enhancing generalisation across heterogeneous services. The training energy consumption increases from 0.72 J to 1.20 J as the number of services increases. However, this increase remains moderate relative to the scale of the system. This is because the NAS service performs architectural search centrally, reducing computation on edge services. Overall, OrchNAS scales effectively, maintaining stable performance while ensuring controlled growth in energy consumption across a large edge environment.

\subsection{Experiment 4: Sensitivity Analysis of OrchNAS}
\label{subsec:sensitivity}
\begin{figure}[t]
\centering
\includegraphics[width=\columnwidth]{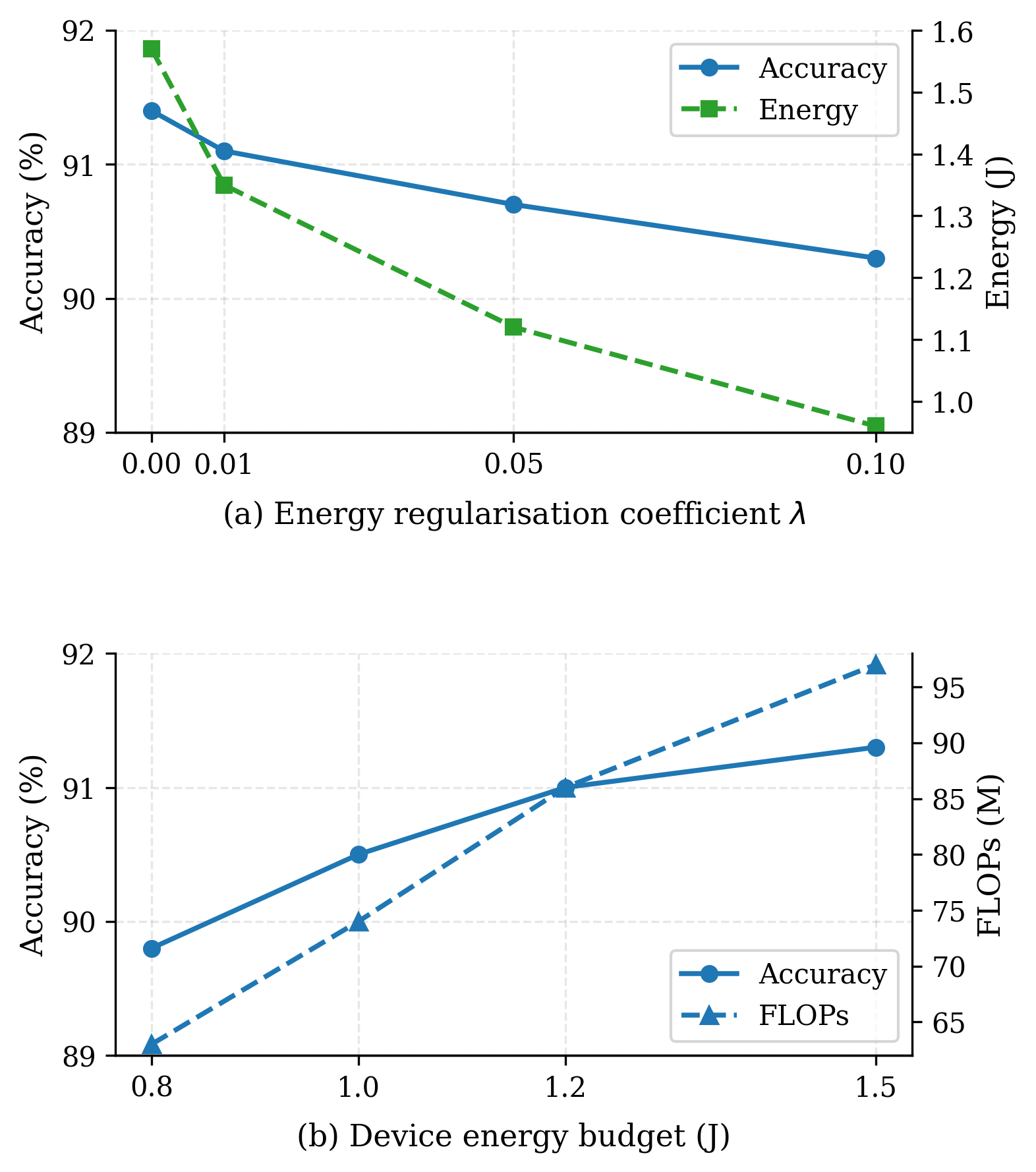}
\caption{Sensitivity analysis of OrchNAS: (a) Effect of Energy Regularisation Coefficient $\lambda$  and (b) Effect of Device Energy Budget}
\label{fig:sensitivity}
\end{figure}

We further evaluate the robustness of OrchNAS under different hyperparameter settings and system configurations. We analyse the impact of the energy regularisation coefficient \(\lambda\), the device energy budget. To ensure a controlled and interpretable analysis, we adopt fixed energy budgets while varying $\lambda \in \{0, 0.01, 0.05, 0.1\}$. This design isolates the effect of energy regularisation on the learned architectures and training behaviour. The results demonstrated that the smaller values of \(\lambda\) favour accuracy-oriented architectures with higher computational and energy cost, whereas increasing \(\lambda\) progressively shifts the search towards more compact and energy-efficient models, with only marginal accuracy degradation. Furthermore, under stricter energy budgets, OrchNAS consistently adapts by selecting smaller subnet architectures that satisfy device-level constraints. This behaviour is important in critical domains such as drone-based surveillance and wearable healthcare, where energy constraints are strict, and models must balance accuracy with efficient resource utilisation. The sensitivity analysis results are shown in Fig.~\ref{fig:sensitivity} (a)(b). These results confirm the robustness of OrchNAS in balancing accuracy and efficiency under varying resource constraints.

\subsection{Discussion}
Our experiments show that OrchNAS improves model accuracy and reduces energy consumption across heterogeneous edge services. The NAS Service orchestration enables efficient exploration of the architecture space, and the energy-aware global architecture search, combined with personalised subnet selection, ensures that service-adaptive constraints are satisfied. This design enables the framework to deliver adaptive, efficient models suitable for real-world edge intelligence scenarios. One limitation of the framework is its focus on energy-aware optimisation, which does not fully capture trade-offs between accuracy, latency, and long-term resource usage. This limitation can lead to suboptimal model selection in scenarios where multiple objectives must be balanced. Future work can address this by incorporating multi-objective optimisation strategies. Another limitation is that personalised architecture search introduces additional computational overhead and limits scalability in large-scale deployments. This can be addressed by adopting global optimisation or group-based adaptation to reduce redundancy and improve scalability.

\section{Conclusion}
In this paper, we propose OrchNAS, an energy-aware personalised federated edge intelligence framework that leverages a NAS Service to automatically design service-adaptive architectures for heterogeneous edge environments. The framework integrates three key components: an energy-efficient global architecture search, an energy-efficient architecture selection mechanism, and a primal–dual optimisation-based personalised approach to enforce strict energy constraints. Together, these components enable effective coordination between global knowledge sharing and service-level adaptation. Experimental results on datasets show that OrchNAS consistently improves accuracy and significantly reduces energy consumption compared to existing PFL with NAS-based approaches. Moreover, the framework achieves high feasibility rates under dynamic service constraints and scales effectively with an increasing number of edge services, demonstrating its practicality for real-world deployments. Future work will explore multi-objective optimisation strategies and more efficient collaborative search mechanisms to further improve scalability and deployment efficiency in complex edge intelligence scenarios.

\end{document}